\documentclass[displayweek]{sid}
\usepackage{subcaption}
\usepackage{booktabs}
\usepackage{enumitem}
\setlist{noitemsep,topsep=2pt,parsep=0pt,partopsep=0pt}

\begin{document}
% stops page numbering throughout document
\pagenumbering{gobble} 

% Tighter spacing
% \setlength{\textfloatsep}{8pt plus 1pt minus 1pt}
% \setlength{\floatsep}{8pt plus 1pt minus 1pt}
% \setlength{\intextsep}{8pt plus 1pt minus 1pt}
% \setlength{\abovedisplayskip}{5pt}
% \setlength{\belowdisplayskip}{5pt}
% \setlength{\abovecaptionskip}{4pt}
% \setlength{\belowcaptionskip}{2pt}

\title{
XStreamVGGT: Extremely Memory-Efficient Streaming Vision Geometry Grounded Transformer with KV Cache Compression }
% XStreamVGGT: Joint Quantization and Pruning for KV Cache Compression in Streaming Geometry-Grounded Transformers
% XStreamVGGT: Extremely Memory-Efficient KV Cache Compression for Streaming Geometry-Grounded Transformers

\author{Zunhai Su$^{1,*}$, Weihao Ye$^{2,*}$, Hansen Feng$^{1}$, Keyu Fan$^{1}$, Jing Zhang$^{3}$, \\Dahai Yu$^{4}$, Zhengwu Liu$^{5}$, Ngai Wong$^{5}$\\
\affiliations $^{1}$Shenzhen International Graduate School, Tsinghua University \\ $^{2}$Institute of Artificial Intelligence, Xiamen University \\
$^{3}$China Star Optoelectronics Technology
$^{4}$TCL Corporate Research (HK) Co., Ltd.\\
$^{5}$Department of Electrical and Electronic Engineering, The University of Hong Kong  
$^{*}$ \textit{Equal contribution}
}

% \author{Weihao Ye$^{1,*}$, Zunhai Su$^{2,*}$, Hansen Feng$^{2}$, Keyu Fan$^{2}$, Yuxin Cheng$^{3}$, Ran Liao$^{2}$, Ngai Wong$^{3}$\\
% \affiliations $^{1}$Institute of Artificial Intelligence, Xiamen University \\ $^{2}$Shenzhen International Graduate School, Tsinghua University \\
% $^{3}$Department of Electrical and Electronic Engineering, The University of Hong Kong 
% $^{*}$\textit{Equal contribution}
% }

\date{}

\maketitle

% force 9pt Times New Roman font for the document body
\small
\begin{abstract}
Learning-based 3D visual geometry models have benefited substantially from large-scale transformers. 
Among these, StreamVGGT leverages frame-wise causal attention for strong streaming reconstruction, but suffers from unbounded KV cache growth, leading to escalating memory consumption and inference latency as input frames accumulate.
We propose XStreamVGGT, a tuning-free approach that systematically compresses the KV cache through joint pruning and quantization, enabling extremely memory-efficient streaming inference.
Specifically, redundant KVs originating from multi-view inputs are pruned through efficient token importance identification, enabling a fixed memory budget.
Leveraging the unique distribution of KV tensors, we incorporate KV quantization to further reduce memory consumption.
Extensive evaluations show that XStreamVGGT achieves mostly negligible performance degradation while substantially reducing memory usage by 4.42$\times$ and accelerating inference by 5.48$\times$, enabling scalable and practical streaming 3D applications.
% Learning-based 3D reconstruction has advanced through transformers like StreamVGGT, which enables streaming reconstruction via frame-wise causal attention but suffers unbounded KV cache growth. 
% We propose XStreamVGGT, a tuning-free approach compressing KV cache through joint pruning and quantization. 
% By pruning redundant tokens and applying distribution-aware quantization, XStreamVGGT achieves 4.42$\times$ memory reduction and 5.48$\times$ speedup with mostly negligible performance degradation, enabling scalable streaming applications.
The code is available at \url{https://github.com/ywh187/XStreamVGGT/}.

\end{abstract}

\begin{keywords}
3D Computer Vision, Streaming Visual Geometry Models, KV Cache Optimization
\end{keywords}

\section{Introduction}
% StreamVGGT and its problems
Recovering 3D geometric structure from image sequences has long been a fundamental problem in computer vision, forming the basis of numerous real-world applications~\cite{samavati2023deep}.
Recently, learning-based feed-forward models have reshaped this field, shifting the paradigm from traditional methods based on geometric priors and iterative optimization to end-to-end transformers that demonstrate strong robustness and cross-dataset generalization \cite{wang2025vggt,wang2024dust3r,wang2025continuous}.
As a milestone in this evolution, the Visual Geometry-Grounded Transformer (VGGT) unifies multiple 3D vision tasks within a single framework and consistently outperforms task-specific approaches across a range of tasks, including dense depth estimation, point map regression, and camera pose prediction \cite{wang2025vggt}.
To support streaming applications, StreamVGGT further replaces the global attention with frame-wise causal attention \cite{zhuo2025streaming}, following a design philosophy analogous to autoregressive large language models (LLMs).

\begin{figure}[t]
\centering
%\vspace{-5mm}
\includegraphics[width=1\graphicswidth]{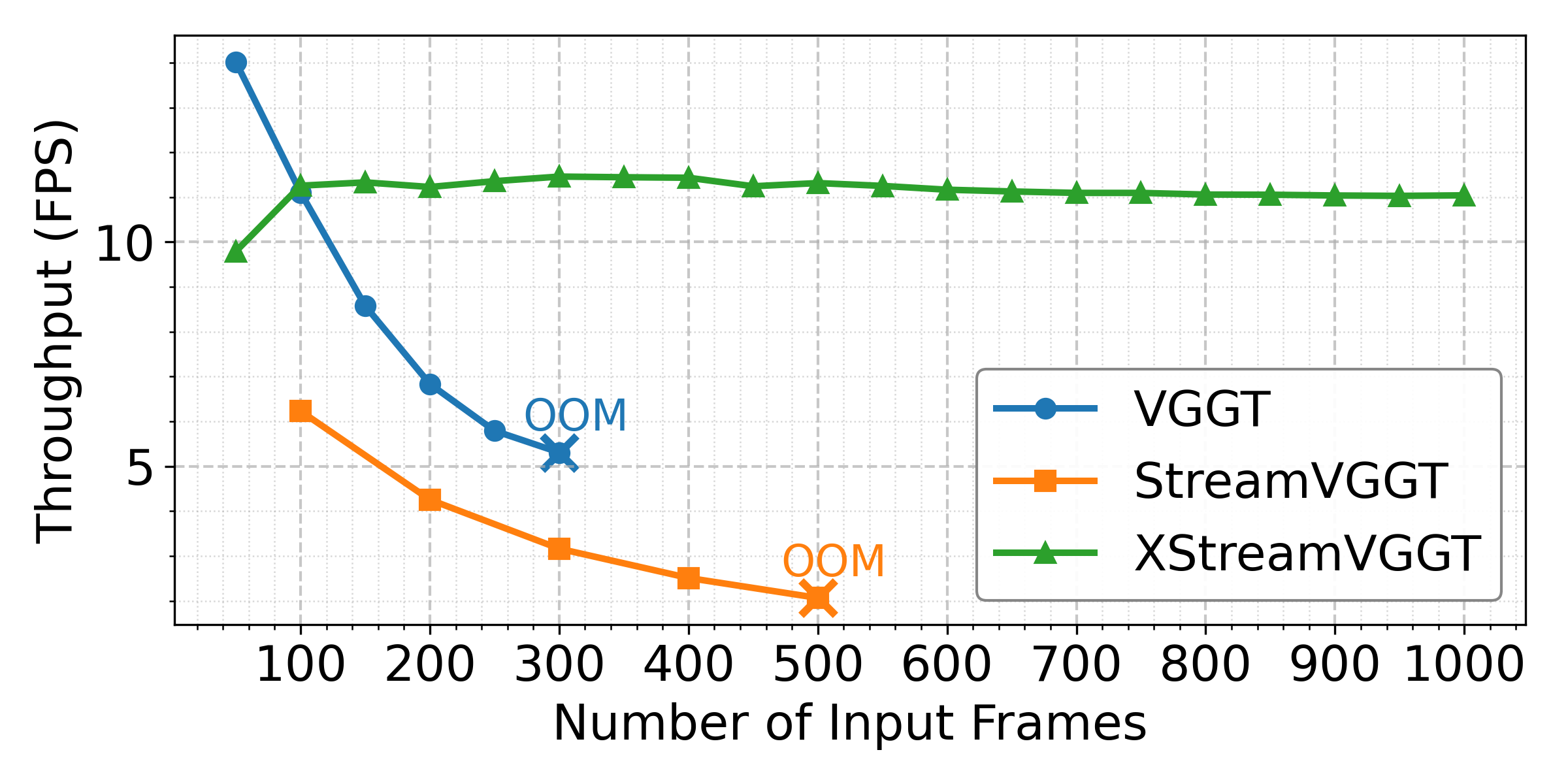}
%\vspace{-2mm}
\caption{Efficiency analysis on a single 80GB A100 GPU.
As the number of input frames increases, StreamVGGT and VGGT exhibit rapid FPS degradation and quickly encounter out-of-memory (OOM) errors, whereas XStreamVGGT consistently maintains higher FPS without OOM.}
%\vspace{-3mm}
\label{fig:fps}
\end{figure}

At the core of StreamVGGT lies its reliance on the Key and Value (KV) cache of previous input frames as an explicit and persistent memory mechanism.
However, as streaming inference proceeds, the KV cache grows linearly with the number of input frames and inevitably becomes unbounded.
As illustrated in Figure \ref{fig:fps}, this unrestrained growth leads to rapidly escalating memory consumption and inference latency, posing a severe bottleneck to scalable and real-world deployment.
% in this work
We propose XStreamVGGT, a tuning-free approach that systematically compresses the KV cache through joint pruning and quantization for extremely memory-efficient streaming inference.
Specifically, XStreamVGGT first eliminates redundant KV cache from multi-view inputs via an efficient token importance identification strategy, pruning the cache to a bounded budget while preserving the first-frame KVs as geometric references~\cite{wang2025vggt}.
Moreover, our analysis reveals pronounced channel-wise outliers in the Key tensors, whereas the Value tensors exhibit much weaker outlier behavior.
Motivated by these observations, we develop a per-channel Key and per-token Value quantization scheme that is seamlessly integrated with pruning to further reduce the memory footprint.
\begin{figure*}[t]
\centering
%\vspace{-7mm}
\includegraphics[width=1\linewidth]{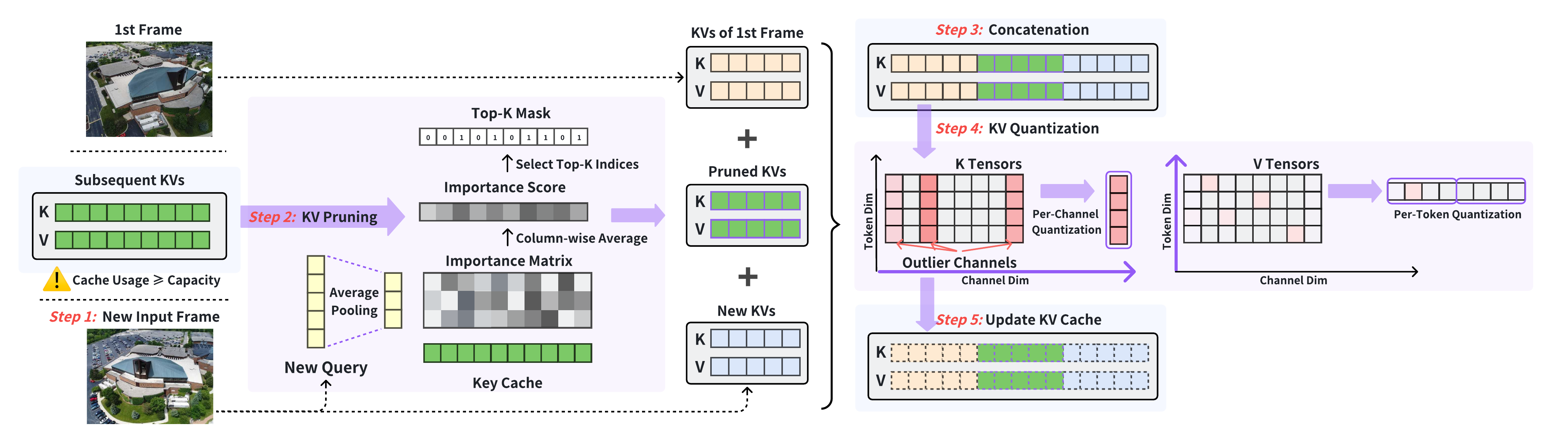}
%\vspace{-2mm}
\caption{
Overview of XStreamVGGT.
Upon receiving a new input frame (Step~1), Queries from the global attention layer are aggregated via average pooling to form a compact representation, which is then matched against the Key to estimate token importance (Step~2). 
Guided by these Key-derived importance scores, low-importance historical KV pairs are selectively pruned, while KVs from the first frame are explicitly retained to preserve geometric consistency. 
The remaining high-importance KVs are concatenated with the first-frame KVs and the newly generated KVs from the current frame (Step~3). 
Finally, the updated KV cache is compressed via quantization yielding a compact cache for subsequent updates (Steps~4–5).
}
%\vspace{-5mm}
\label{fig:method}
\end{figure*}
% contribuations
Our key contributions can be summarized as follows:
\begin{itemize}
    %\vspace{-2mm}
    \item We introduce XStreamVGGT, the first method that systematically combines KV cache pruning and quantization to enable highly memory-efficient streaming 3D applications.
    
    \item Extensive evaluations on video depth estimation, 3D reconstruction, and camera pose estimation demonstrate that XStreamVGGT achieves mostly negligible performance degradation while reducing memory usage by 4.42$\times$ and accelerating inference by 5.48$\times$.
%\vspace{-1mm}
\end{itemize}

\section{Related Work}

\subsection{Learning-Based 3D Reconstruction.}
% 3D reconstruction has undergone a paradigm shift: from methods heavily reliant on handcrafted priors to fully data-driven, learning-based frameworks. 
Leveraging neural networks to implicitly encode scene priors, recent learning-based approaches have achieved markedly improved robustness and generalization, setting new benchmarks across diverse 3D vision tasks \cite{wang2025vggt}. 
Progress was marked by DUSt3R \cite{wang2024dust3r}, which directly regresses view-consistent 3D pointmaps from two RGB images without requiring camera calibration. 
CUT3R \cite{wang2025continuous} introduces a stateful recurrent framework that incrementally updates a scene representation from a stream of images. 
% TTT3R further extends CUT3R by adopting a Test-Time Training strategy, dynamically adjusting memory updates based on alignment confidence between past states and new observations to improve length generalization. 
VGGT~\cite{wang2025vggt} represents a key milestone, scaling this philosophy to a 1.2B-parameter transformer that jointly predicts multiple 3D features, achieving state-of-the-art performance across a range of 3D vision tasks.
StreamVGGT \cite{zhuo2025streaming} extends VGGT to support online streaming reconstruction. 
However, its unbounded KV cache poses a critical limitation, hindering practical deployment in real-world streaming scenarios.

\subsection{KV Cache Compression}
KV cache enables efficient inference by avoiding redundant recomputation of past KVs. 
As context lengths increase, the KV cache can grow into a substantial memory bottleneck, emphasizing the need for effective compression \cite{liu2024kivi}. 
KV pruning methods reduce memory and computation by discarding past KVs of low estimated importance, typically guided by attention scores or heuristic saliency \cite{ye2025fit,wu2024accelerating}. 
% Such approaches reveal that much of the cached KVs are redundant, allowing significant cache reduction with little effect on model performance~\cite{ye2025fit,wu2024accelerating}.
KV quantization methods, by representing high-precision floating-point tensors in low-bit formats, enable compact KV cache storage~\cite{liu2024kivi, su2025rotatekv,su2025kvsink}. 
Nevertheless, existing methods are designed for LLMs, and the compression of KV caches in 3D vision models remains largely unexplored.

\section{Methodology}
% In this section, we present XStreamVGGT, with an overview illustrated in Figure~\ref{fig:method}. 
% We first review the preliminaries of StreamVGGT, followed by our approach to eliminating multi-view redundancy through KV cache pruning. 
% We then analyze the distributional properties of KV tensors in StreamVGGT, which motivate the design of an effective KV quantization scheme.
\subsection{Preliminaries}
StreamVGGT is a streaming visual geometry transformer that processes video frames online and produces per-frame geometric outputs without reprocessing the entire history~\cite{zhuo2025streaming}.
Given an incoming RGB frame $I_t \in \mathbb{R}^{3 \times H \times W}$ at time step $t$, the model first converts it into a sequence of visual tokens $F_t \in \mathbb{R}^{N \times C}$ via a patch embedding network, where $N$ denotes the number of image patches and $C$ is the embedding dimension.
In addition to the patch tokens, StreamVGGT prepends a camera token $g_t \in \mathbb{R}^{1 \times C}$ and $R$ register tokens $r_t \in \mathbb{R}^{R \times C}$.
The resulting token sequence is expressed as
$
X_t = [\, g_t;\, r_t;\, F_t \,] \in \mathbb{R}^{(1 + R + N) \times C}.
$
$X_t$ is then fed into a spatio-temporal transformer encoder comprising $L$ layers with an alternating-attention design.
% Specifically, each layer applies frame-wise spatial self-attention within the current frame, followed by temporal causal attention that aggregates information from past frames under a causal constraint.

In each layer $\ell$, prior to temporal attention computation, the model maintains a KV cache that stores KVs from all previous frames,
% \begin{equation}
% \mathcal{C}^{(\ell)}_{t-1}
% =
% \bigl\{ K^{(\ell)}_{1:t-1},\; V^{(\ell)}_{1:t-1} \bigr\},
% \end{equation}
% where $K^{(\ell)}_{\tau}, V^{(\ell)}_{\tau} \in \mathbb{R}^{(1 + R + N) \times C}$ denote the Key and Value tensors corresponding to frame $\tau$.
For the current frame at time step $t$, only $Q^{(\ell)}_{t}$, $K^{(\ell)}_{t}$, and $V^{(\ell)}_{t}$ are newly computed.
Temporal attention is then formulated as
\begin{equation}
%\vspace{-3mm}
\mathrm{Attn}\!\left(
Q^{(\ell)}_{t},\;
\bigl[ K^{(\ell)}_{1:t-1},\; K^{(\ell)}_{t} \bigr],\;
\bigl[ V^{(\ell)}_{1:t-1},\; V^{(\ell)}_{t} \bigr]
\right).
\end{equation}
% where a causal mask is applied to prevent access to future frames.
After attention computation, the newly generated $K^{(\ell)}_{t}$ and $V^{(\ell)}_{t}$ are appended to the KV cache.

\subsection{Eliminating Multi-View Redundancy through KV Cache Pruning }
% The temporal KV cache grows linearly with the number of processed frames, rapidly exceeding memory constraints in long-term streaming scenarios.
% Moreover, as illustrated in Fig X, multi-view inputs exhibit substantial redundancy, which can be exploited to compress the KV cache.
We propose a query-guided KV cache pruning mechanism to eliminate multi-view redundancy while retaining the most informative historical tokens within a fixed cache length $\mathcal{L}_{\text{max}}$.
Given the Query $Q^{(\ell)}_t \in \mathbb{R}^{(1+R+N) \times C}$ of the current frame, we first separate the special tokens (camera token and register tokens) $Q^{(\ell)}_{t, \text{special}}$ from the ordinary patch tokens $Q^{(\ell)}_{t, \text{normal}}$. 
We group normal tokens with length $g$ and average each group to obtain a compact representation, followed by averaging over all attention heads to form the pooled query:
\begin{figure}[t]
    \centering  
    %\vspace{-3mm}
    \begin{subfigure}{0.48\linewidth}
        \centering
    \includegraphics[width=\linewidth]{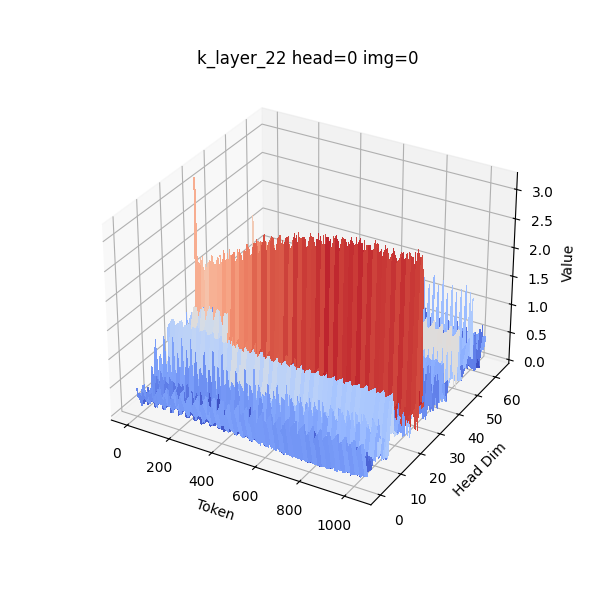}
    %\vspace{-10mm}
    \caption{Key tensor.}
    \end{subfigure}
    \begin{subfigure}{0.48\linewidth}
        \centering
    \includegraphics[width=\linewidth]{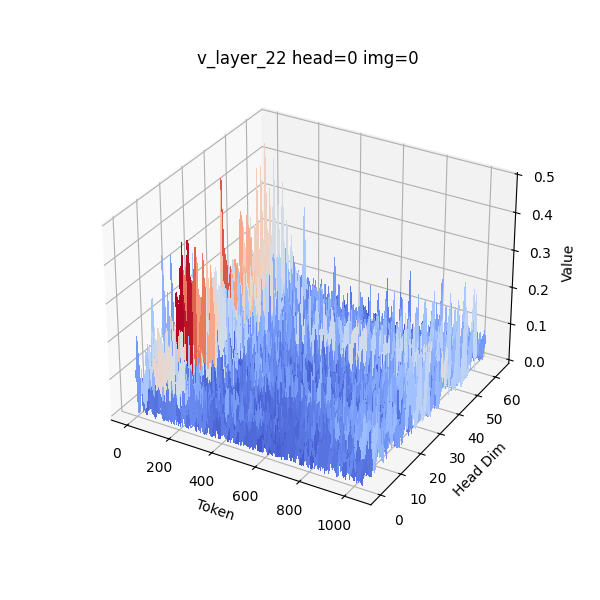}
    %\vspace{-10mm}
    \caption{Value tensor.}
    \end{subfigure}
\caption{Magnitude distributions of the Key and Value.
The Key exhibits pronounced channel-wise outliers, with a small number of channels having larger magnitudes than the rest.}
    % %\vspace{-3mm}
\label{fig:kv tensor}
\end{figure}
\begin{equation}
%\vspace{-3mm}
Q^{(\ell)}_{t, \text{pooled}} =
\text{concat}\Big(
Q^{(\ell)}_{t, \text{special}},
\; \text{GroupAvg}(Q^{(\ell)}_{t, \text{normal}}, g)
\Big).
\end{equation}
% %\vspace{-3mm}
\begin{equation}
%\vspace{-2mm}
\bar{Q}^{(\ell)}_t =
\frac{1}{H} \sum_{h=1}^{H} Q^{(\ell)}_{t, \text{pooled}}[h]
\in \mathbb{R}^{N_{\text{pooled}} \times C}.
\end{equation}
Here, $\text{GroupAvg}(\cdot, g)$ reshapes normal tokens into non-overlapping groups of size $g$ and averages along the token dimension, and $N_{\text{pooled}}$ denotes the number of tokens after grouping.

Pruning is triggered when the cache length $T$ exceeds the budget $\mathcal{L}_{\text{max}}$. 
Tokens from the first frame and the current frame are always preserved as a stable geometric reference and up-to-date visual evidence, respectively.
Let $T_{\text{first}}$ and $T_{\text{current}}$ denote the numbers of tokens from the first and current frames.
The remaining middle segment,
$
T_{\text{prunable}} = T - T_{\text{first}} - T_{\text{current}},
$
is subject to pruning.
For these tokens, we compute a head-averaged key summary:
\begin{equation}
%\vspace{-2mm}
\bar{K}^{(\ell)}_{\text{prunable}}
=
\frac{1}{H}
\sum_{h=1}^{H}
K^{(\ell)}_{1:t-1}
\bigl[h,\; T_{\text{first}} : T - T_{\text{current}} \bigr]
\in \mathbb{R}^{T_{\text{prunable}} \times C}.
\end{equation}

Token importance scores are computed via the inner product between the pooled queries and the prunable keys, followed by averaging along the query dimension:
\begin{equation}
%\vspace{-2mm}
S^{(\ell)}_{\text{matrix}}
=
\bar{Q}^{(\ell)}_t
\left(\bar{K}^{(\ell)}_{\text{prunable}}\right)^{\!\top}
\in
\mathbb{R}^{N_{\text{pooled}} \times T_{\text{prunable}}},
\end{equation}
\begin{equation}
S^{(\ell)}
=
\frac{1}{N_{\text{pooled}}}
\sum_{i=1}^{N_{\text{pooled}}}
S^{(\ell)}_{\text{matrix}}[i, :]
\in
\mathbb{R}^{T_{\text{prunable}}}.
\end{equation}
This group-wise query pooling reduces both noise and computational cost while preserving the dominant information demand of the current frame.
Notably, this procedure is fully compatible with highly optimized attention kernels such as FlashAttention \cite{dao2023flashattention}.
Based on the resulting importance scores, the top-$k$ tokens are selected from the prunable middle region, where
\(
k = \max\bigl(0,\; \mathcal{L}_{\text{max}} - T_{\text{first}} - T_{\text{current}} \bigr).
\)
Let $\mathcal{I}_{\text{middle}}$ denote the indices of the selected tokens in the middle region.
The final set of retained tokens is then given by
\(
\mathcal{I}_{\text{keep}}
=
\{1, \dots, T_{\text{first}}\}
\;\cup\;
\mathcal{I}_{\text{middle}}
\;\cup\;
\{T - T_{\text{current}} + 1, \dots, T\}.
\)
% Finally, the pruned KV cache is constructed by gathering along the token dimension,
% \begin{equation}
% \tilde{K}^{(\ell)}_{1:t-1}
% =
% K^{(\ell)}_{1:t-1}[\mathcal{I}_{\text{keep}}],
% \qquad
% \tilde{V}^{(\ell)}_{1:t-1}
% =
% V^{(\ell)}_{1:t-1}[\mathcal{I}_{\text{keep}}],
% \end{equation}
% which is then concatenated with the Key and Value tensors of the current frame to form the updated cache $\mathcal{C}^{(\ell)}_t$.
This strategy retains only the most informative historical tokens while always preserving the first and current frames, ensuring geometric consistency and temporal continuity.

\begin{table}[t]
  \centering
    %\vspace{-5mm}
    \caption{3D reconstruction comparison on 7-Scenes dataset.}
    %\vspace{-1mm}
    \resizebox{1\columnwidth}{!}{%
    \begin{tabular}{@{}c|cccccc@{}}
    \toprule
     & \multicolumn{6}{c}{7 Scenes} \\ \cmidrule(l){2-7} 
     & \multicolumn{2}{c}{Acc↓} & \multicolumn{2}{c}{Comp↓} & \multicolumn{2}{c}{NC↑} \\ \cmidrule(l){2-7} 
    Method & Mean & Med. & Mean & Med. & Mean & Med. \\ \midrule
    % CUT3R & 0.120 & 0.047 & 0.157 & 0.030 & 0.724 & 0.831 \\
    StreamVGGT & 0.132 & 0.058 & 0.116 & 0.042 & 0.749 & 0.863 \\
    XStreamVGGT & 0.142 & 0.068 & 0.125 & 0.048 & 0.734 & 0.848 \\ \bottomrule
    \end{tabular}%
    }
    %\vspace{-2mm}
  \label{tab:recons-1}%
\end{table}%
\begin{table}[t]
  \centering
    \caption{3D reconstruction comparison on NRGBD dataset.}
    %\vspace{-1mm}
    \resizebox{1\columnwidth}{!}{%
    \begin{tabular}{@{}c|cccccc@{}}
    \toprule
     & \multicolumn{6}{c}{NRGBD} \\ \cmidrule(l){2-7} 
     & \multicolumn{2}{c}{Acc↓} & \multicolumn{2}{c}{Comp↓} & \multicolumn{2}{c}{NC↑} \\ \cmidrule(l){2-7} 
    Method & Mean & Med. & Mean & Med. & Mean & Med. \\ \midrule
    % CUT3R & 0.100 & 0.036 & 0.078 & 0.031 & 0.834 & 0.970 \\
    StreamVGGT & 0.085 & 0.044 & 0.079 & 0.038 & 0.862 & 0.986 \\
    XStreamVGGT & 0.085 & 0.049 & 0.075 & 0.038 & 0.850 & 0.986 \\ \bottomrule
    \end{tabular}%
    }
    %\vspace{-2mm}
  \label{tab:recons-2}%
\end{table}%
\begin{table}[t]
  \centering
  %\vspace{-5mm}
  \caption{Camera pose estimation.}
    %\vspace{-1mm}
    \resizebox{1\columnwidth}{!}{%
    \begin{tabular}{@{}c|ccc|ccc@{}}
    \toprule
     & \multicolumn{3}{c|}{TUM} & \multicolumn{3}{c}{ScanNet} \\ \cmidrule(l){2-7} 
    Method & ATE↓ & RPE trans↓ & RPE rot↓ & ATE↓ & RPE trans↓ & RPE rot↓ \\ \midrule
    % CUT3R & 0.047 & 0.147 & 0.451 & 0.094 & 0.022 & 0.629 \\
    StreamVGGT & 0.062 & 0.033 & 3.208 & 0.160 & 0.057 & 3.688 \\
    XStreamVGGT & 0.068 & 0.035 & 3.184 & 0.171 & 0.061 & 3.837 \\ \bottomrule
    \end{tabular}%
    }
    %\vspace{-2mm}
  \label{tab:camera pose}%
\end{table}%
\begin{table}[t]
    \centering
    \caption{Video depth estimation.}
    %\vspace{-1mm}
    \resizebox{1\linewidth}{!}{%
    \begin{tabular}{@{}c|cc|cc|cc@{}}
    \toprule
     & \multicolumn{2}{c|}{Sintel} & \multicolumn{2}{c|}{Bonn} & \multicolumn{2}{c}{KITTI} \\ \cmidrule(l){2-7} 
    Method & Abs Rel↓ & $\delta<$1.25↑ & Abs Rel↓ & $\delta<$1.25↑ & Abs Rel↓ & $\delta<$1.25↑ \\ \midrule
    % CUT3R & 0.432 & 46.7 & 0.078 & 93.7 & 0.122 & 87.6 \\
    % StreamVGGT & 0.328 & 65.8 & 0.058 & 97.2 & 0.173 & 72.2 \\
    % XStreamVGGT & 0.341 & 61.9 & 0.073 & 94.3 & 0.206 & 63.9 \\ 
    StreamVGGT & 0.328 & 65.8 & 0.058 & 95.9 & 0.094 & 94.4 \\
    XStreamVGGT & 0.341 & 61.9 & 0.077 & 97.1 & 0.098 & 94.3 \\ 
    \bottomrule
    \end{tabular}%
    %\vspace{-2mm}
    }
  \label{tab:video_depth}%
\end{table}%
% \begin{table}[]
% \resizebox{\columnwidth}{!}{%
% \begin{tabular}{@{}c|cc|cc|cc@{}}
% \toprule
%  & \multicolumn{2}{c|}{Sintel} & \multicolumn{2}{c|}{Bonn} & \multicolumn{2}{c}{KITTI} \\ \cmidrule(l){2-7} 
% Method & Abs Rel↓ & $\delta<$1.25↑ & Abs Rel↓ & $\delta<$1.25↑ & Abs Rel↓ & $\delta<$1.25↑ \\ \midrule
% StreamVGGT & 0.328 & 65.8 & 0.058 & 97.2 & 0.173 & 72.2 \\
% XStreamVGGT & 0.341 & 61.9 & 0.073 & 94.3 & 0.206 & 63.9 \\ \midrule
% StreamVGGT-clip to 30 frames & - & - & 0.058 & 95.9 & 0.094 & 94.4 \\
% XStreamVGGT-clip to 30 frames & - & - & 0.077 & 97.1 & 0.098 & 94.3 \\ \midrule
% StreamVGGT-clip to 50 frames & - & - & 0.059 & 96.7 & 0.123 & 86.0 \\
% XStreamVGGT-clip to 50 frames & - & - & 0.082 & 96.8 & 0.145 & 79.7 \\ \bottomrule
% \end{tabular}%
% }
% \end{table}
\subsection{Quantization Guided by KV Distribution Characteristics}
In this work, we adopt the widely used asymmetric uniform quantization scheme \cite{liu2024kivi}, which is parameterized by a scale factor $s$, a zero-point $z$, and a bit-width $b$.
Given a floating-point tensor $x \in \mathbb{R}^{d}$, its quantized representation $\hat{x}$ is computed as
\begin{equation}
%\vspace{-2mm}
\hat{x}
=
\operatorname{clamp}
\left(
\left\lfloor \frac{x}{s} \right\rceil + z,\;
0,\;
2^{b}-1
\right),
\end{equation}
with the scale factor and zero-point defined as
\begin{equation}
%\vspace{-2mm}
s
=
\frac{x_{\max} - x_{\min}}{2^{b} - 1},
\qquad
z
=
\left\lfloor -\frac{x_{\min}}{s} \right\rceil.
\end{equation}
Here, $\lfloor \cdot \rceil$ denotes rounding to the nearest integer, and $\operatorname{clamp}(\cdot)$ restricts values to the valid range.
The scale factor $s$ determines the quantization step size, while the zero-point $z$ ensures that the real-valued zero is exactly representable within the integer domain.

After establishing the basic quantization strategy, we next determine the appropriate quantization granularity (i.e., per-tensor, per-token, or per-channel).
A principled understanding of the distributional properties of KV tensors in StreamVGGT is crucial for achieving robust and accurate quantization.
% Our analysis reveals, for the first time, that the KV cache in StreamVGGT exhibits distributional characteristics that fundamentally differ from those observed in LLMs.
As illustrated in Fig.~\ref{fig:kv tensor}, the Key tensors in StreamVGGT exhibit pronounced channel-wise outliers, whereas such behavior is much less evident in the Value tensors.
When applying standard per-tensor or per-token quantization, these outliers dominate the dynamic range, severely inflating the quantization scale and drastically reducing the effective precision, which results in substantial performance degradation \cite{nagel2021white}.

Guided by these observations, we design a per-channel Key and per-token Value quantization scheme that explicitly accounts for channel-wise outliers.
In contrast to LLMs, where the KV cache typically grows by a single token at each decoding step, StreamVGGT processes inputs in a frame-wise manner.
This results in the KV cache expanding by a large number of patch tokens per time step, making StreamVGGT inherently well suited for per-channel Key quantization schemes that operate across multiple tokens.
As illustrated in Fig.~\ref{fig:method}, we tightly couple quantization with pruning.
Specifically, quantization is applied to the final KV cache, $\tilde{K}^{(\ell)}_{1:t-1}$ and $\tilde{V}^{(\ell)}_{1:t-1}$, which comprises both the pruned historical KVs and the preserved KVs from the first and current frames.
% We apply per-channel quantization to both the Key and Value along the channel (head-dimension) axis to alleviate the impact of channel-wise outliers commonly observed in KV tensors.
The quantization is defined as
\begin{equation}
%\vspace{-2mm}
\hat{K}_{c} = \mathcal{Q}_{c}\!\left(\tilde{K}_{c}; \, s^{K}_{c}, \, z^{K}_{c}\right),
\qquad
\hat{V}_{t} = \mathcal{Q}_{t}\!\left(\tilde{V}_{t}; \, s^{V}_{t}, \, z^{V}_{t}\right),
\end{equation}
where $c$ and $t$ denote channel and token indices, respectively, 
and $s^{K}_{c}, z^{K}_{c}$ (for Keys) and $s^{V}_{t}, z^{V}_{t}$ (for Values) are the corresponding scale and zero-point parameters.
Here, $\mathcal{Q}_{c}(\cdot)$ and $\mathcal{Q}_{t}(\cdot)$ denote asymmetric uniform quantization along the channel and token dimensions, respectively.
During attention computation, the quantized tensors are dequantized as
\begin{equation}
%\vspace{-2mm}
\tilde{K}_{c} = \mathcal{DQ}_{c}\!\left(\hat{K}_{c}; \, s^{K}_{c}, \, z^{K}_{c}\right),
\qquad
\tilde{V}_{t} = \mathcal{DQ}_{t}\!\left(\hat{V}_{t}; \, s^{V}_{t}, \, z^{V}_{t}\right),
\end{equation}
where $\mathcal{DQ}_{c}(\cdot)$ and $\mathcal{DQ}_{t}(\cdot)$ denote the dequantization operators.

\begin{figure}[t]
\centering
%\vspace{-7mm}
\includegraphics[width=1\graphicswidth]{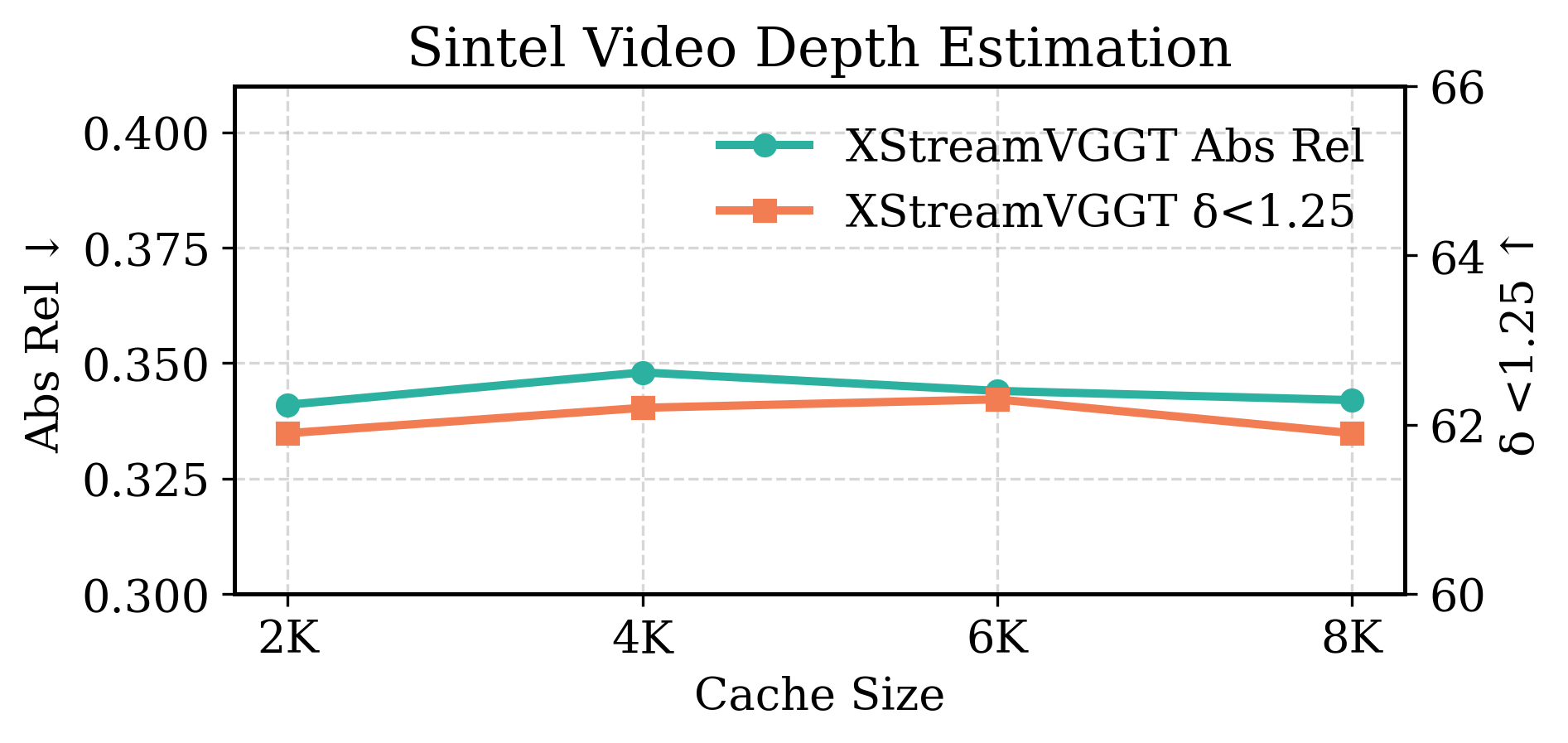}
%\vspace{-2mm}
\caption{Ablation study of cache length.}
\label{fig:cache length}
\end{figure}

\begin{figure}[t]
\centering
%\vspace{-3mm}
\includegraphics[width=0.9\graphicswidth]{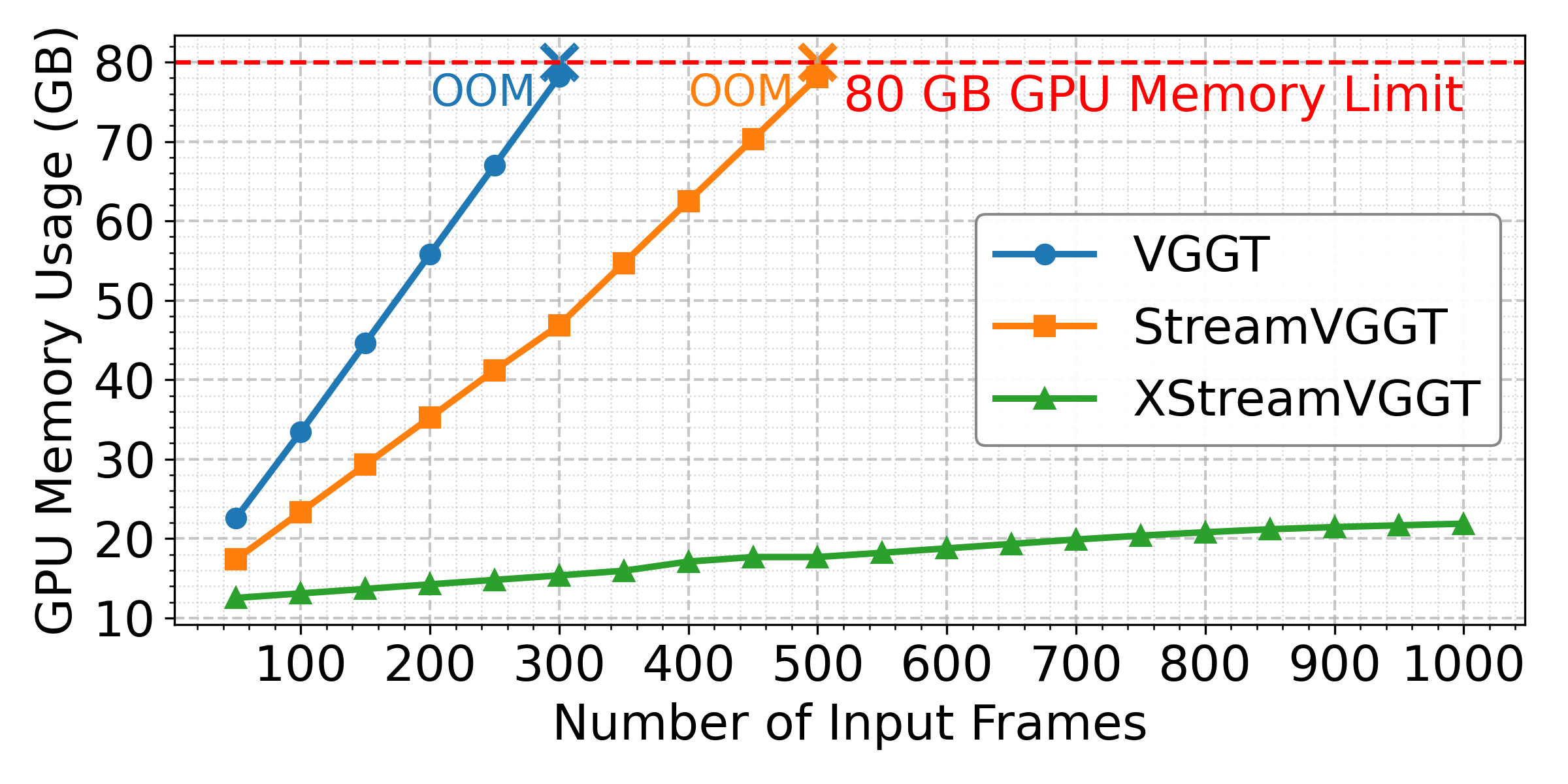}
%\vspace{-2mm}
\caption{Analysis of memory with increasing frame length.}
%\vspace{-2mm}
\label{fig:memory}
\end{figure}

\begin{figure}[t]
    \centering    
    %\vspace{-5mm}
    \begin{subfigure}{0.49\columnwidth}
        \centering
    \includegraphics[width=\linewidth]{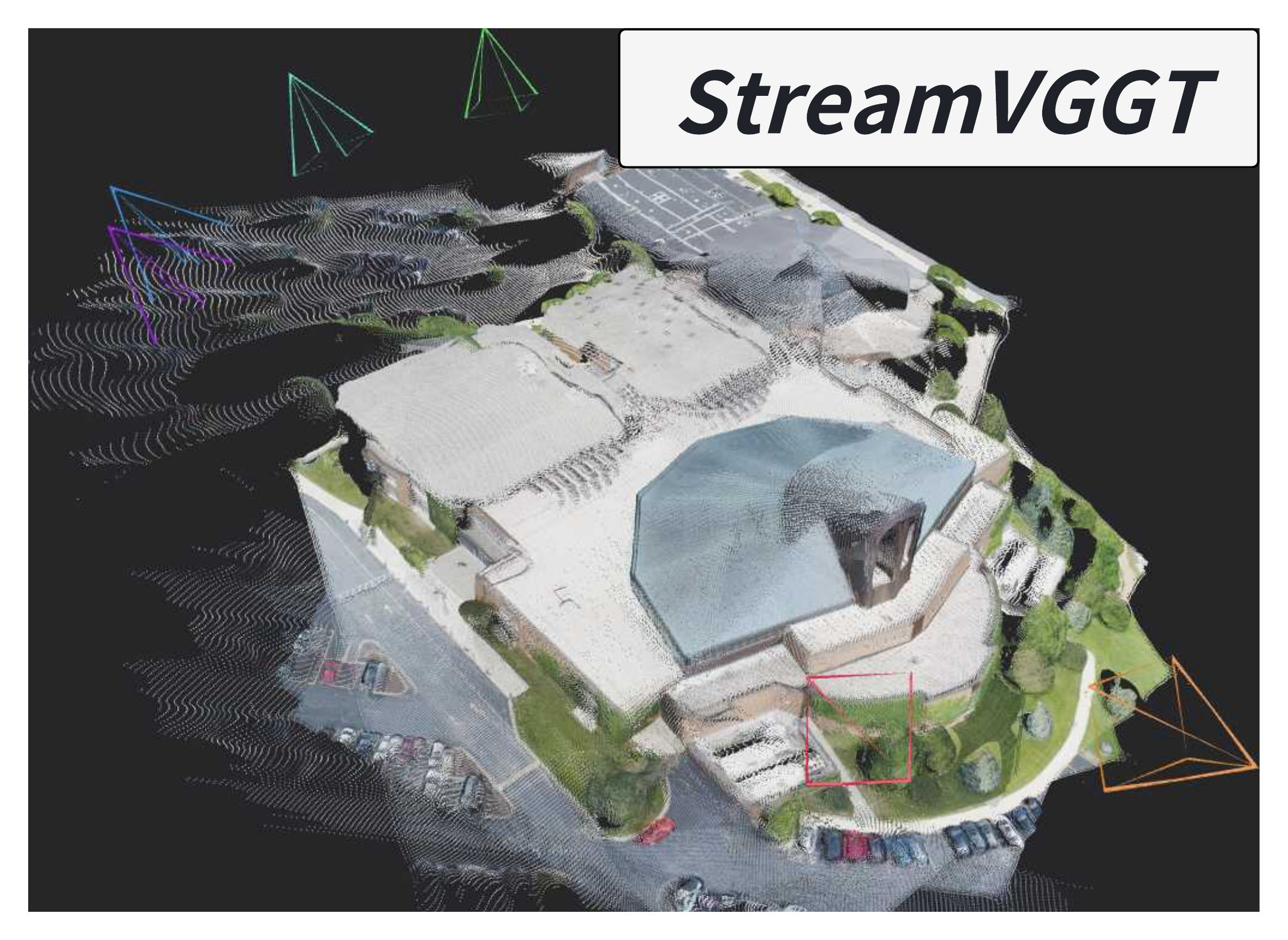}
    % %\vspace{-5mm}
    % \caption{StreamVGGT.}
    \end{subfigure}
    \begin{subfigure}{0.49\columnwidth}
        \centering
    \includegraphics[width=\linewidth]{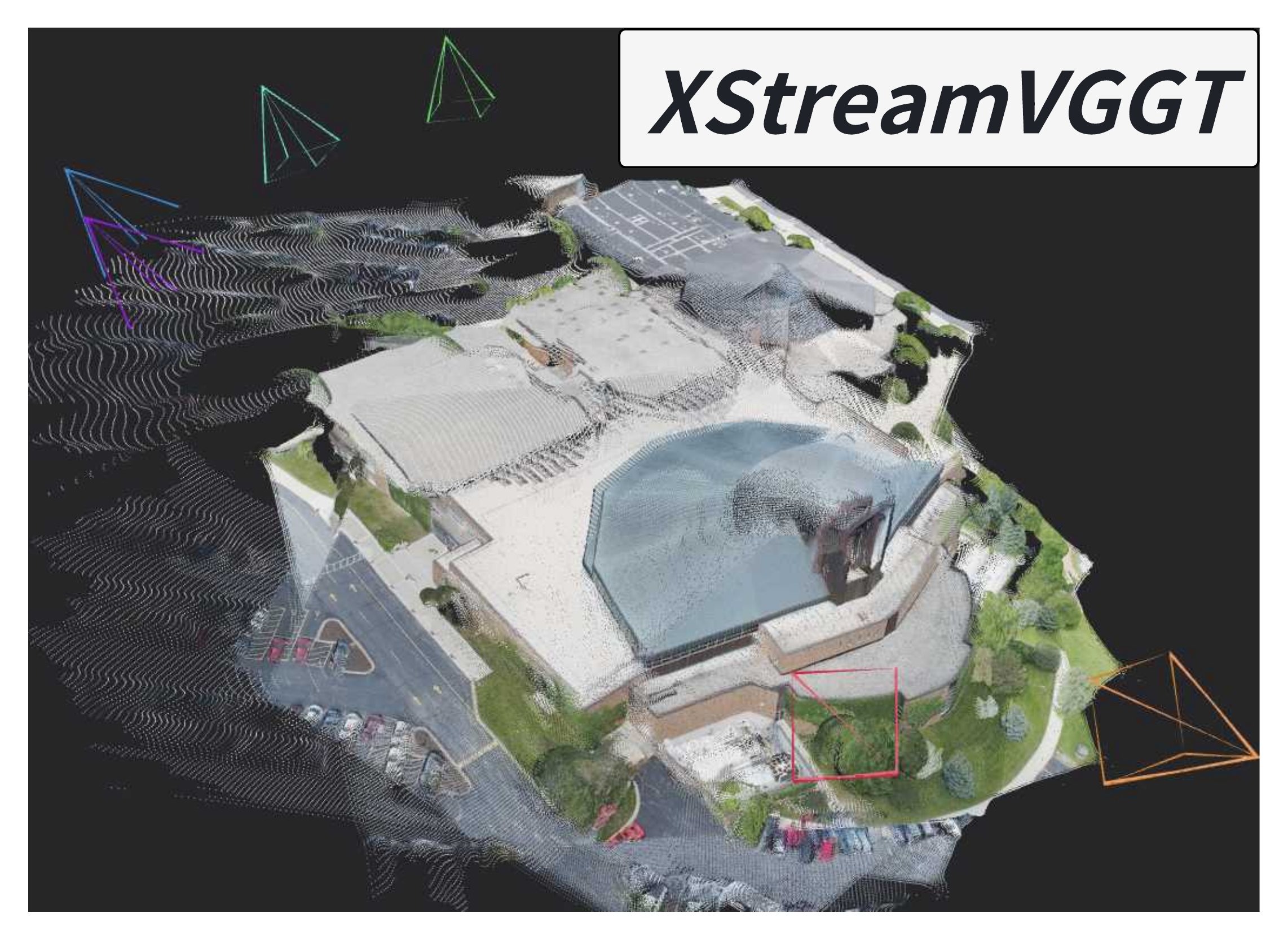}
    % %\vspace{-5mm}
    % \caption{XStreamVGGT.}
    \end{subfigure}
    \begin{subfigure}{0.49\columnwidth}
        \centering
    \includegraphics[width=\linewidth]{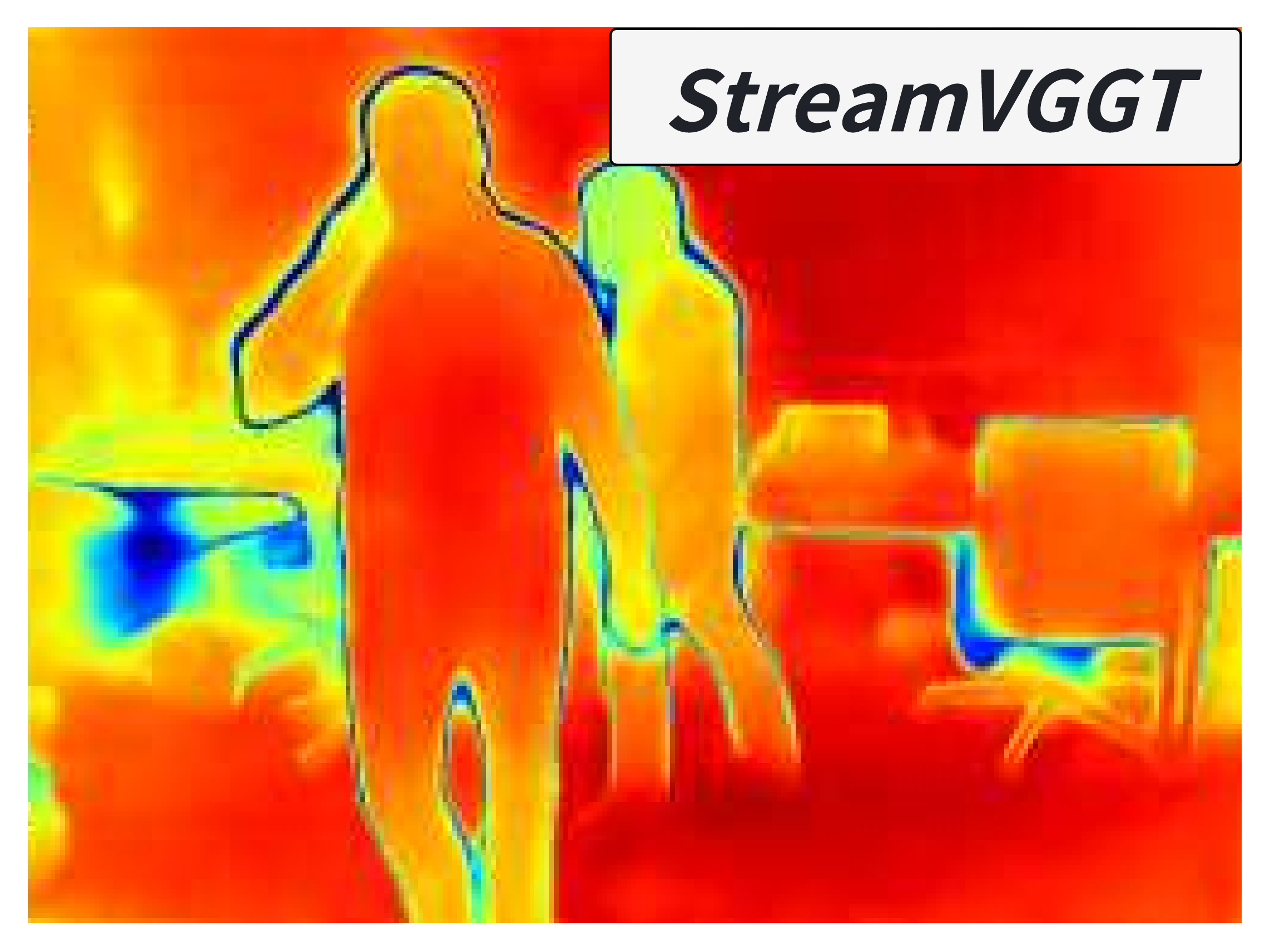}
    % %\vspace{-5mm}
    % \caption{StreamVGGT.}
    \end{subfigure}
    \begin{subfigure}{0.49\columnwidth}
        \centering
    \includegraphics[width=\linewidth]{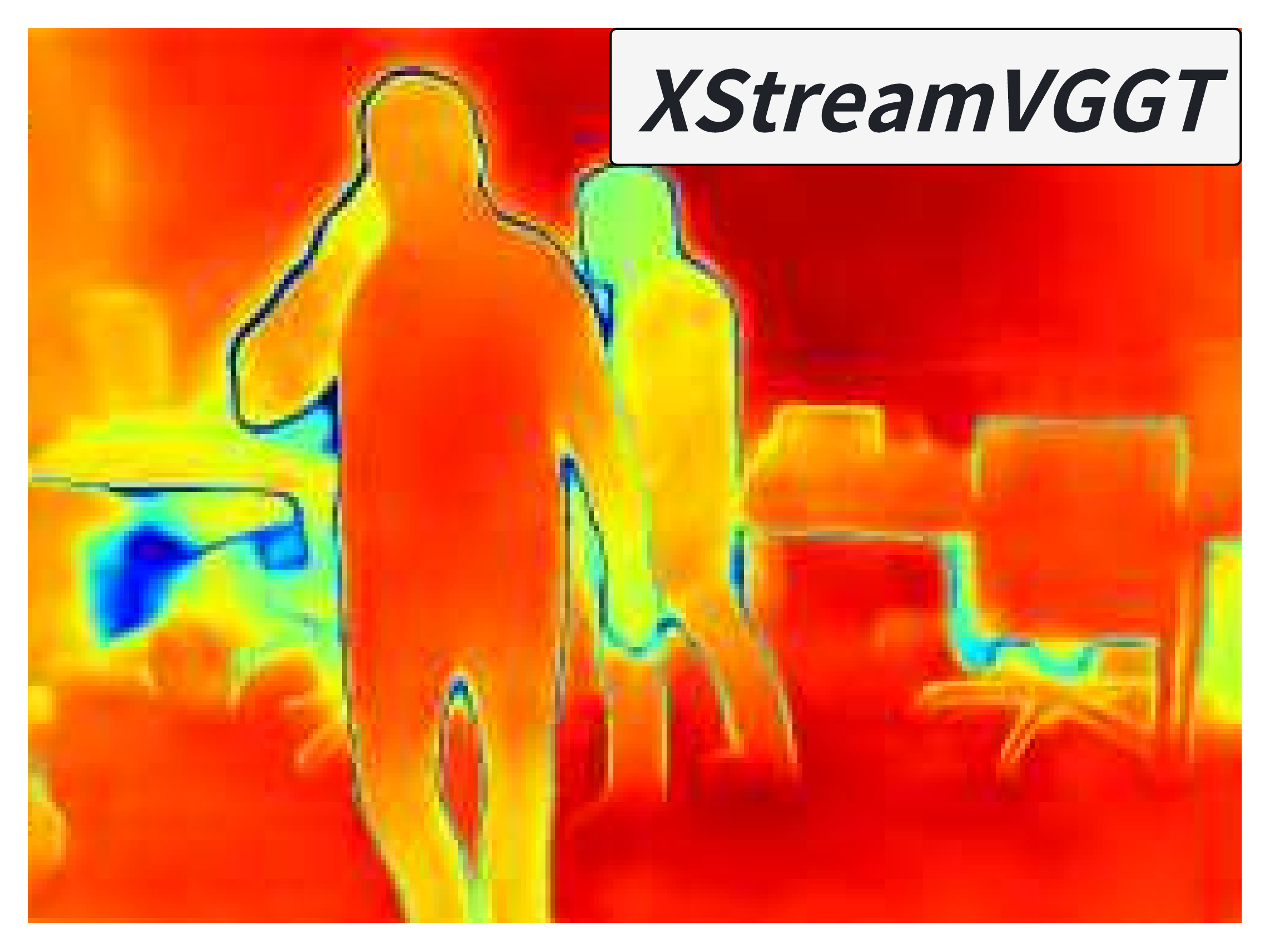}
    % %\vspace{-5mm}
    % \caption{XStreamVGGT.}
    \end{subfigure}
    %\vspace{-1mm}
\caption{Qualitative reconstruction and depth estimation results comparing StreamVGGT and XStreamVGGT.}
    %\vspace{-2mm}
\label{fig:Qualitative results.}
\end{figure}
\section{Experiments}

\subsection{Experimental Details}
We evaluate XStreamVGGT on a variety of 3D tasks, including video depth estimation, 3D reconstruction, and camera pose estimation. 
Following the Point3R protocol \cite{wu2025point3r}, input images are processed with variable aspect ratios and resized such that the maximum edge length is 518 pixels. 
We further assess inference speedup and memory savings to demonstrate the efficiency gains of XStreamVGGT. 
The pooling size is set to 16 and the cache length to 2K to ensure that the evaluations can at least preserve the first and current frames. 
% We also perform an extensive ablation study on the cache length.
KV quantization is performed using KIVI with INT4 and a group size of 64 \cite{liu2024kivi}.
Notably, even with a cache length of only 2K and additional quantization, XStreamVGGT achieves exceptional efficiency while maintaining strong performance, with mostly negligible loss across most tasks compared to the expensive full-KV-length StreamVGGT \cite{zhuo2025streaming}.

\subsection{3D Reconstruction}
The 3D reconstruction experiments were conducted on the 7-Scenes \cite{shotton2013scene} and NRGBD \cite{azinovic2022neural} datasets by measuring the discrepancies between predictions and the corresponding ground-truth point clouds.  
We adopt Accuracy (Acc), Completion (Comp), and Normal Consistency (NC) as evaluation metrics, and benchmark the performance on full-length sequences.  
As shown in Tables \ref{tab:recons-1} and \ref{tab:recons-2}, XStreamVGGT demonstrates strong geometric fidelity, achieving a mean NC score of 0.734 on the 7-Scenes dataset—only a 2\% drop compared with StreamVGGT (0.749).
% On the NRGBD dataset, XStreamVGGT attains a median NC of 0.654, indicating that for the majority of scenes, it can generate 3D point clouds with accurate surface orientations and well-preserved details, achieving nearly identical performance to StreamVGGT.

\subsection{Camera Pose Estimation}
We evaluate camera pose estimation on the TUM Dynamics \cite{sturm2012benchmark} and ScanNet \cite{dai2017scannet} datasets, truncating all sequences to 90 frames.
We report Absolute Translation Error (ATE), Relative Translation Error (RPE\textsubscript{trans}), and Relative Rotation Error (RPE\textsubscript{rot}) as evaluation metrics.
As shown in Table \ref{tab:camera pose}, XStreamVGGT achieves nearly lossless performance across all three metrics compared with StreamVGGT.

\subsection{Video Depth Estimation}
We conduct video depth estimation on the Sintel \cite{butler2012naturalistic}, Bonn \cite{palazzolo2019refusion}, and KITTI \cite{geiger2013vision} datasets, covering a wide range of dynamic and static, indoor and outdoor scenes.  
Absolute relative error (Abs Rel) and $\delta < 1.25$ (the percentage of predicted depths within a 1.25-factor of the ground-truth depth) are adopted as evaluation metrics.  
As shown in Table~\ref{tab:video_depth}, XStreamVGGT incurs mostly negligible performance degradation with a 2K cache length. 
While the relative drop on Bonn appears noticeable due to its exceptionally strong baseline, XStreamVGGT retains very high absolute performance.

\subsection{Ablation Study}
Figure~\ref{fig:cache length} presents an ablation study on cache length on the Sintel dataset, varying across 2K, 4K, 6K, and 8K. 
As shown, performance changes minimally with increasing cache length, indicating significant redundancy among multi-view patch tokens. 
A cache length of 2K maintains strong performance while remaining highly efficient, making it a suitable choice.

\subsection{Efficiency Analysis}
We evaluate the efficiency of long-sequence modeling on the TUM Dynamics and ScanNet datasets, using sequences ranging from 50 to 1000 frames to measure GPU memory usage and inference speed.
As shown in Figures~\ref{fig:fps} and~\ref{fig:memory}, XStreamVGGT achieves substantial efficiency gains, reducing memory usage by 4.42$\times$ and accelerating inference by 5.48$\times$.

\subsection{Qualitative Results}
Figure~\ref{fig:Qualitative results.} presents visual comparisons between StreamVGGT and XStreamVGGT.
XStreamVGGT closely preserves visual quality, producing results consistent with StreamVGGT while achieving significantly higher efficiency.

\section{Conclusion}
We introduce XStreamVGGT, a tuning-free method for memory-efficient streaming inference in vision geometry–grounded transformers. 
By combining KV cache pruning with quantization, it bounds memory growth while preserving model fidelity. 
Extensive experiments demonstrate minimal performance loss alongside substantial reductions in memory footprint and inference latency, enabling scalable 3D streaming applications.
Future work will explore adaptive cache budgets that dynamically adjust based on scene complexity and motion characteristics.

\section{Acknowledgements}
This work was supported in part by the Research Grants Council of the Hong Kong Special Administrative Region Government through the TRS project T45-701/22-R and GRF project 17203224, and the TCL Corporate Research (Hong Kong) Co., Limited.

% \bibliographystyle{vancouver}
% \bibliography{XStreamVGGT/XStreamVGGT}

\end{document}